\documentclass[journal]{IEEEtran}

\usepackage{url}
\usepackage{amsmath}
\usepackage{amsfonts}
\usepackage{graphicx}
\usepackage{multirow}
\usepackage{color}
\usepackage{cancel}
\usepackage{fancyhdr}

\fancypagestyle{firstpage}{%
  \lfoot{\scriptsize{\textcopyright 2019 IEEE. Paper published in IEEE Transactions on Neural Networks and Learning Systems (DOI: 10.1109/TNNLS.2019.2890970). Personal use of this material is permitted.  Permission from IEEE must be obtained for all other uses, in any current or future media, including reprinting/republishing this material for advertising or promotional purposes, creating new collective works, for resale or redistribution to servers or lists, or reuse of any copyrighted component of this work in other works.}}
  \pagenumbering{gobble}
}

\begin{document}
%

\title{Natural Language Statistical Features of LSTM-generated Texts}

%

\author{Marco~Lippi$^*$,
	Marcelo~A~Montemurro$^*$,
        Mirko~Degli~Esposti,
	and~Giampaolo~Cristadoro
\thanks{$^*$These two authors contributed equally. Marco Lippi is with the
Department of Sciences and Methods for Engineering, University of Modena and
Reggio Emilia, email: marco.lippi@unimore.it. Marcelo A Montemurro is with the
Division of Neuroscience and Experimental Psychology, University of Manchester,
Manchester, UK, email: m.montemurro@manchester.ac.uk. Mirko Degli Esposti is
with the Department of Computer Science and Engineering, University of Bologna,
email: mirko.degliesposti@unibo.it. Giampaolo Cristadoro is with the Department
of Mathematics and Applications, University of Milano Bicocca, email:
giampaolo.cristadoro@unimib.it.}
}

%
%

\maketitle
\thispagestyle{firstpage}

\begin{abstract}
Long Short-Term Memory (LSTM) networks have recently shown remarkable
performance in several tasks dealing with natural language generation, such as
image captioning or poetry composition. Yet, only few works
have analyzed text generated by LSTMs in order to quantitatively evaluate to
which extent such artificial texts resemble those generated by humans. We
compared the statistical structure of LSTM-generated language to that of written natural language,
and to those produced by Markov models of various orders. In particular, we characterized the statistical
structure of language by assessing word-frequency statistics, long-range correlations,
and entropy measures. Our main finding is that while both LSTM and Markov-generated
texts can exhibit features similar
to real ones in their word-frequency statistics and entropy measures,
LSTM-texts are shown to reproduce long-range correlations at scales comparable
to those found in natural language.
Moreover, for LSTM networks a temperature-like parameter controlling the generation process shows an optimal value---for which the produced texts are closest to real language---consistent across the different statistical features investigated.
\end{abstract}


\section{Introduction}
\label{sec:introduction}

Building artificial systems capable of mimicking human creativity has long been one of the aims of Artificial Intelligence (AI)~\cite{Boden1998}. In this work, we focus on the problem of Natural Language Generation (NLG), which encompasses the capability of machines to synthesize text in a way that resembles spoken or written language typically employed by humans~\cite{Reiter2000}.
This research field has recently known a period of great excitement, mostly due to the huge development in the area of deep learning~\cite{LeCun2015}, whose methods and algorithms have certainly contributed to move significant steps forward.

Deep learning techniques have in fact produced stunning results in a variety of different research fields and application domains, and one of the major successes has been that of generative models~\cite{Bengio2009}. In the area of NLG, many studies have been dedicated to specific, focused applications, such as image and video captioning~\cite{Karpathy2015,Venugopalan2015}, poem synthesis~\cite{Zhang2014} or lyric generation~\cite{Potash2015}. In all these cases, the considered generated texts are relatively short (captions, poems, lyrics) and correlations between words rarely span across several sentences. The scenario totally changes when we consider longer texts, such as novels. Natural language has been widely studied within this context, and notoriously it shows statistical properties in the distribution of terms, as well as long-range correlations between words~\cite{Ebeling1995, Montemurro2002, Altmann2012}.
In comparison to short texts such as captions, this is a much more challenging setting to imitate for machines.

In this work, we aim to provide an extensive empirical evaluation of texts generated with Long-Short Term Memory (LSTM) networks, one of the most widely used deep learning models for NLG. Our goal is to quantitatively assess whether LSTM texts do share some similarities with natural language that is commonly produced by humans.
To this aim, we trained an LSTM network with a corpus that consists of a collection of novels by Charles Dickens. Such network is trained to predict the next character of a given text, and thus it can be employed to iteratively generate a document of any desired length. The setting was adopted in several works (e.g., see~\cite{Graves2013,Karpathy2015understanding} and citations therein).

In our experimental framework we evaluated several different aspects of machine-generated texts, comparing them against the statistics of real language and Markov-generated samples. First, we analyzed fundamental linguistic properties typically shown by texts, such as Zipf's~\cite{Zipf1935} and Heaps'~\cite{Heaps1978} laws for words. Second, we studied whether the generated texts presented long-range correlations, which are commonly encountered in human-generated texts, but difficult to reproduce for machines. As a third point, we compared the entropy of the generated texts with the one of the original corpus. We then moved our analysis to a higher level, by carrying out a preliminary study looking at characteristics dealing with the style and quality of the generated texts: in particular, we analyzed the degree of creativity and plagiarism of the artificial texts with respect to the dataset on which the LSTM was trained, by looking at longest common subsequences. We also assessed whether an authorship attribution algorithm would capture some analogy between the generated text and the original one, in terms of author's style.

Surprisingly, very few studies have been dedicated to a thorough analysis and to a quantitative evaluation of the similarities between texts created by machines and texts created by humans. Karpathy et al.~\cite{Karpathy2015understanding} also provide an experimental analysis of LSTM networks trained for character-by-character text generation, but they focused their study on a qualitative evaluation of the cell activations within the neural architecture: for example, they looked for open and closed parentheses or quotes, that typically span a few tens or hundreds of characters. Their claim that the LSTM model is capable of capturing long-range dependencies is thus only supported by such qualitative evidence, without giving a deep insight in the characteristics of the generated documents. Lin and Tegmark~\cite{Lin2017} compared natural language texts with those generated by Markov models and LSTMs, exploiting metrics coming from information theory. Their analysis shows that LSTMs are capable of capturing correlations that Markov models instead fail to represent, yet the range of correlations they consider is still quite limited (up to 1,000 characters). Conversely, Takahashi and Tanaka-Ishii~\cite{TakaTana2017} reported that LSTM language models have limitation in reproducing such long-range correlations if measured with a method based on clustering properties of rare words; note however that their analysis is still limited to a range of $\sim$1,000 words, and the corpus they employ for training is much smaller than the one used in our experiments.
Ghodsi and DeNero~\cite{Ghodsi2016} instead, analyzed some statistical properties of text generated by a Recurrent Neural Network Language Model (RNNLM), in particular focusing on the length of sentences, the vocabulary distribution, and the distribution of specific grammatical elements, such as pronouns.
In a complementary study, Lake and Baroni~\cite{LakeBaroni2017}  focused on short-scale structures instead of the overall statistical properties of pseudo-texts. They showed that RNNs are able to exploit systematic compositionality, and thus also to reproduce, for example, abstract grammatical generalizations.

The remainder of the paper is organized as follows. Section~\ref{sec:lstm} will describe the LSTM model used in our experiments. Section~\ref{sec:methods} will present the statistical methods employed for the quantitative evaluation of the artificial text properties. Section~\ref{sec:corpora} will describe the corpora used to train our model, and Section~\ref{sec:experiments} will report and discuss the experimental results. Section~\ref{sec:conclusions} will finally conclude the paper, also pointing for future research directions.

\section{Long Short-Term Memory Networks}
\label{sec:lstm}

Long Short-Term Memory networks (LSTMs) are recurrent neural networks (RNNs) that have been first developed at the end of the 90s, achieving remarkable results in applications dealing with input sequences~\cite{Hochreiter1997}. Such model was specifically designed to address the issue of vanishing gradients, that greatly limited the applicability of standard RNNs~\cite{Bengio1994}. Within the ``deep learning revolution'' that Artificial Intelligence has been undergoing in the last decade, LSTMs have regained popularity, being now widely used in a huge number of research and industrial applications, including automatic machine translation, speech recognition, text-to-speech generation (e.g., see~\cite{Greff2017} and references therein).

\subsection{General framework}

RNNs allow to process sequences of arbitrary lengths, by exploiting $L$ hidden layers $h_t^\ell$, with $\ell = \{1, \ldots, L\}$ whose cells are functions not only of the layer input $x_t$, but also of the hidden layer at the previous time step: $h_t^\ell = f(x_t, h_{t-1}^\ell)$. RNNs are typically trained with Backpropagation Through Time (BPTT)~\cite{Werbos1990}, by unfolding the recurrent structure into a sort of temporal chain through which the gradient is propagated, up to a certain number $K$ of time steps. Unfortunately, this method suffers from the well-known problem of vanishing or exploding gradients~\cite{Hochreiter1997,Bengio1994}, which makes plain RNNs scarcely used in practice. LSTMs overcome this issue by exploiting a more complex hidden cell, namely a \textit{memory cell}, and non-linear gating units, that control the information flows into and out of the cell.

\begin{figure}
\begin{center}
\includegraphics[width=0.95\columnwidth]{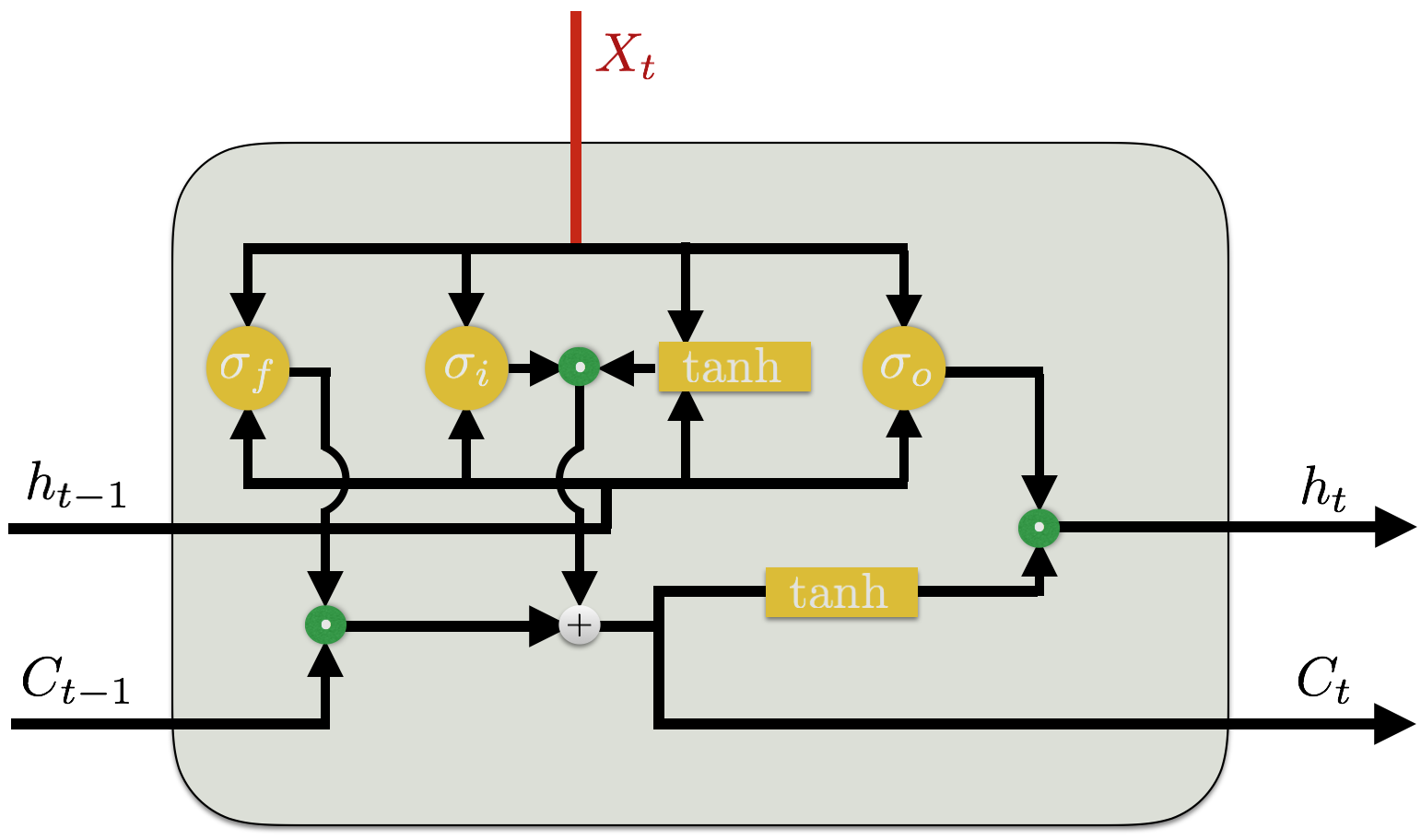}
\caption{Depiction of LSTM cell. $\sigma_f, \sigma_i, \sigma_o$ are typically the sigmoid function. Recurrent connections $h_t$ and $C_t$ propagate information through time.\label{fig:lstm}}
\end{center}
\end{figure}

Basically, the LSTM cells are capable of maintaining their state over time, of forgetting what they have learned, and also of allowing novel information in.
An example of such a cell is depicted in Figure~\ref{fig:lstm}. The model is based on the concept of \textit{cell status} at time $t$, namely $C_t$, which depends on three \textit{gates}: an \textit{input} gate $i_t$ that can let new information into the cell state, a \textit{forget} gate $f_t$ that can modulate how much information is forgotten from the previous state, and an \textit{output} gate $o_t$ that controls how much information is transferred to the upper layers.
The following equations describe the behaviour of an LSTM layer (we drop the layer index $\ell$ in order to simplify the notation):
\begin{align}
  f_t &= \sigma_f (W_f x_t + U_f h_{t-1} + b_f)\\
  i_t &= \sigma_i (W_i x_t + U_i h_{t-1} + b_i)\\
  o_t &= \sigma_o (W_o x_t + U_o h_{t-1} + b_o)\\
  C_t &= f_t \odot C_{t-1} + i_t \odot \tanh (W_c x_t + U_c h_{t-1} + b_c)\\
  h_t &= o_t \odot \tanh(C_t)
\end{align}
where all $\sigma_f, \sigma_i, \sigma_o$ are typically the sigmoid function, $\odot$ indicates the Hadamard (or element-wise) product, and $W$, $U$, $b$ represent the model parameters that have to be learned. As a form of regularization, dropout is nowadays typically employed in deep neural networks~\cite{Srivastava2014}. Dropout simply consists in randomly dropping a percentage $(1-p)$ of connections between neurons during training, while multiplying by $p$ every weight in the network at testing time. In recurrent architectures, though, this general framework does not work well in practice, but dropout can still be successfully employed, if applied only to inter-layer connections and not to recurrent ones~\cite{Zaremba2014}. This is how we employed dropout in our model.

\subsection{LSTM for text generation}

The most widely employed LSTM architecture for text generation is based on character-level sentence modeling. Basically, the input of the network consists in $M$ characters, that correspond to a fixed-size portion of text, whereas the number of output neurons is the total number $S$ of possible symbols in the text, each neuron corresponding to one of such characters. The output layer consists in a softmax layer, so that each symbol has an associated probability, and all such probabilities sum up to 1. A hard way of generating texts is to pick the character with the highest probability as the next one in the sequence, and to feed it back into the network input. A soft alternative (which is the one used in practice) allows to sample the next character in the sequence from the probability distribution of the output cells: in this way the output of the network is not deterministic, given the same initial input sequence. With such an iterated procedure, texts of any length can be generated. The hidden states of the cells keep track of the ``memory'' of the network, so as to exploit also information not directly encoded in the input any more.
This generation phase can be controlled by a parameter $T$, usually named temperature. Different temperature values can be tuned in order to obtain a smoother or sharper softmax distribution from which characters are generated: in particular, the final softmax layer in the LSTM computes the probability of each symbol $j$ as follows:
\begin{equation}
  P_j = \frac{\exp(\frac{y_j}{T})}{\sum_{i=1}^S \exp(\frac{y_i}{T})}
\end{equation}
where $y_i$ are the output values for each symbol that are fed into softmax. Large $T$ values lead towards a uniform distribution of symbol probabilities, whereas when $T$ tends to zero, the distribution is skewed towards the most probable symbol.

Such a model is trained in a classic supervised learning setting, where the input training corpus is fed to the network, using as target the true (known) next character, as it appears in the corpus. If the cell states are not reset as subsequent input windows are presented to the network, long-range dependencies can in principle be captured by the model.

Note that, in principle, the same task could be modeled at a word level, thus training the LSTM network to predict the next \textit{word} in the text, rather than the next character. Although this solution may appear a more appropriate way for modeling the problem, it has two main limitations: (i) the number of possible output classes of the network would become huge, being the total number of distinct words in the input corpus, leading to a much more difficult optimization problem, which would likely require a larger number of training examples; (ii) the set of possible output words would be limited to those appearing in the training corpus, which could be enough in many cases, but would limit the creativity of the network.

\section{Methods}
\label{sec:methods}

We now present the methods we employ in order to quantitatively evaluate the characteristics of the artificially generated texts with respect to the original, human-generated texts.

\subsection{Zipf's and Heaps' laws for words}
\label{sec:ZipfHeap}

Natural languages show remarkable statistical properties in their word statistics.
The two best known examples are Zipf's \cite{Zipf1935} and Heaps' \cite{Heaps1978} laws in language,
which refer to universal features related to word frequencies.

Zipf's law states that if the word frequencies of any sufficiently long text are arranged in decreasing order,
there is a power-law relationship between the frequency and the corresponding ranking order of each word. More explicitly,
if we denote the rank of a word by $r$, the Zipf's law states the following relationship between the rank and the
frequency of a word at that rank position $f(r)$, as follows:

\begin{equation}
 f(r)=A r^{-\beta} .
\end{equation}

This relationship is roughly the same for all human languages, the exponent $\beta$ taking values close to 1.

Heaps' law states that the number of the different words (i.e the size of the vocabulary) after seeing $t$ consecutive words in a text,
obeys approximately the following relationship:

\begin{equation}
n(t)=B {t^{\nu}}
\end{equation}\,
with exponent $\nu$ typically taking values smaller than 1 \cite{Gerlach2013}.

\subsection{Long-Range Correlations}
\label{sec:correlations}

Linguistic laws on the scaling of word frequencies, like Zipf's and Heaps', do
not reveal any statistical structure that depends directly on word order.
Zipf's rank-frequency distribution would be unaltered after a random
text shuffling, and similarly for the average behaviour
predicted by Heaps' law. Statistical measures that
capture structure at the sequence level in texts involve correlations  and
spectral analysis \cite{Voss1975}.

Correlations in language are known to be of the power-law
type~\cite{Ebeling1995, Montemurro2002, Altmann2012}, decaying as $C(\tau)
\propto \tau^{-\gamma}$, where $\tau$ is the distance between
symbols -- e.g. words or characters.
It is
possible to characterize the structure of long-range correlations using the
method of Detrended Fluctuation
Analysis (DFA) \cite{Peng1992, Buldyrev1995}.  The first step in the method
involves the
mapping of the symbol sequence onto a numerical time-series, by assigning a
number
to each basic symbol in the sequence. In order to preserve the maximum of
structure
from the text sources, we performed the mapping at character level---including
all punctuation signs, numbers, capital letters, and accented forms. The
procedure
to assign a number to each character followed similar lines to the method
employed in
\cite{Montemurro2002}, where in the present case each
character is replaced by its rank. Thus, the most frequent character is
assigned
the number 1, the second most frequent the number 2, and so on. For a sequence
of length $N$, the character at position $t$ in the time series, with $t\in
\mathbb{N}$, can be represented by the number $x(t)$, and the following
random-walk can be constructed:
\begin{equation}
 X(t)=\sum_{i=1}^t (x(i)-\langle x \rangle)
\end{equation}
where $\langle x \rangle$ represents the mean of the times series. The
time series $X(t)$
is then split into windows of length $L$, and in each of those windows the
corresponding stretch
of the series is fitted by a straight line by means of least squares. These
linear fits represent the
local trend within each of the windows of length $L$. The sequence of length-$L$ trends
can
be concatenated in a piecewise manner
defining a piecewise linear function $Y_L(t)$. Then we compute the average
fluctuations at scale $L$, that is the deviations from the trend, defined as
follows:
\begin{equation}
 F(L)=\left(\frac{1}{N}\sum_{t=1}^N \left(X(t)-Y_L(t)\right)^2 \right)^{\frac{1}{2}} \, .
\end{equation}

The nature of the correlations present in the original time series can be
evaluated by observing the dependence of $F(L)$ on $L$.
In particular, the growth of fluctuation with the scale $L$ will
be given as $F(L)\sim L^\alpha$, with $0<\alpha<1$. In the case of an
uncorrelated or short-range correlated
time series $x_t$, we have $F(L)\sim L^{\frac{1}{2}}$.
However, in the presence of long-range correlations of the
power-law type in the original time series $x_t$, the fluctuation
exponent $\alpha$ will differ from $1/2$~\cite{Buldyrev1995}, with
$\alpha>1/2$ for persistent (positive) correlations and $\alpha<1/2$ for
anti-persistent (negative) correlations.

\subsection{Entropy and KL-divergence estimation}
\label{sec:entropy}

The entropy of a symbolic sequence can be interpreted as a quantification of
the
degree of predictability in the sequence. A high level of
predictability of consecutive values in a sequence implies a low level
of surprise about future
symbols, which is linked to a low entropy. On the other hand, a sequence with
a high degree of randomness will be
characterized by a high level of surprise in the identity of future symbols,
and result in a high value for the entropy.  Therefore, the determination of the
entropy
of language serves as a quantification of its degree of order.  Early attempts
to determine the
entropy of language were based on the close link between entropy and
predictability
\cite{Shannon1951}. However, the estimation of the entropy from long sequences
of written text, requires
the estimation of block probabilities, which poses a
serious computational challenge, due to the presence of long-range
correlations in language.
The required sample size needed for an accurate estimation of the block
probabilities grows exponentially
with the length of the block, thus quickly rendering insufficient any available
amount of text. This difficulty can be overcome through the
link between entropy and predictability mentioned above. The degree of
predictability in a sequence
determines how much it can be compressed by a lossless compression method.
Sequences with high predictability can be compressed more than sequences with a
higher degree of randomness.
In particular, it can be shown that under general assumptions of stationarity and
ergodicity, the entropy rate of a stochastic source is a lower bound to the
length per symbol of any encoding of it \cite{Cover2006}. Hence, the entropy
of
symbolic sequences can be estimated by means of efficient lossless
compression algorithms~\cite{Lempel1976, Wyner1989,  Schurmann1996}. We estimated the entropy of long character sequences using
an implementation of the algorithm proposed by Lempel and
Ziv~\cite{Lempel1976, Ziv1977,Ziv1978}, which
relies on the estimation of redundancy by looking for matches between future and past
substrings
in a symbolic sequence. Implementations of entropy estimation algorithms based on these methods have proved to work well
for symbolic sequences even in the presence of long range correlations as those
found in language~\cite{Schurmann1996,Kontoyiannis1998}.

The Kullback-Leibler (KL) divergence $D(P|Q)$ is a measure of relative entropy
between
two probability distributions $P$ and $Q$~\cite{Cover2006}. When $P\equiv Q$
then the KL-divergence is zero, but it takes positive values when $P\ne Q$. It
can be
shown that the $D(P|Q)$ is a measure of the extra numbers of bits that are
required to encode typical sequences with
the distribution $P$, when using a code based on $Q$ \cite{Cover2006}. This
interpretation suggests that $D(P|Q)$ can also be estimated
using compression algorithms in which one signal is compressed using past
sequences in the second signal. More specifically, the KL divergence $D(P|Q)$ can
be written
as~\cite{Cover2006}
\begin{equation}
D(P|Q)=H_P(Q)-H(P)
\end{equation}
where $H(P)$ is the entropy of the distribution $P$ and $H_P(Q)$ is
the cross-entropy between $P$ and $Q$. Let us assume that two text sequences
produced by the stochastic source with probability distribution
$P$ are represented by $X=\{x_t\}_{t=1}^N$, and those produced by $Q$ as
$Z=\{z_t\}_{t=1}^N$. Then, for notational succinctness let us write the
information
quantities explicitly in terms of the generated sequences, therefore
representing the KL-divergence between $P$ and $Q$ as $D(X|Z)$. With this
notation,
we have $D(X|Z)=H_X(Z)-H(X)$. A compression-based algorithm
proposed in \cite{Ziv1993} permits to compute the cross-entropy $H_X(Z)$ based
on the symbolic sequences $X$ and $Z$. Then, the KL-divergence
is obtained by subtracting the entropy of the sequence $X$ from the
cross-entropy $H_X(Y)$.  The KL divergence is a non-symmetric quantity and in
order
to have a distance-like measure between character sequences $X$ and $Y$, we
defined a symmetrized divergence as $D_s(X,Y)=(D(X|Y)+D(Y|X))/2$.

Another measure that is strictly related to entropy, and that is widely used
for the evaluation of artificial texts, is perplexity~\cite{Jelinek1977}, which
can be computed as the geometric mean of the inverse probability for each
predicted word in the a document~\cite{Bengio2003,Vinyals2015}, where
probabilities are typically estimated on a language model trained on a larger
corpus.

\subsection{Creativity and Authorship Attribution}
\label{sec:creativity}

Providing a quantitative method able to address the {\it creativity} of a given algorithm for artificial texts generation is a complex task. In this paper we consider two distinct yet strictly intertwined aspects: we aim to measure at what extent the algorithm is capable of capturing the stylistic traits of a given author, while, at the same time, avoiding to perform just a plagiarism of the training corpus.

Measuring the Longest Common Subsequence (LCS) is one of the simplest way to implement the idea of quantitatively measuring plagiarism: given the $k$-th character $x_k$ of the artificial text, we denote by $L_k$ the length of the longest subsequence starting at $x_k$ that is also contained in (thus, plagiarized from) the training corpus.
Different statistics of the set of all $L_k$ can be used to quantify how various algorithms are able to reproduce some stylist traits of a given corpus while generating innovative texts, not written before. Here we adopt the simplest one and consider the maximum over the whole artificial text: $\bar{L}=\max_k L_k$ (see~\cite{Papadopoulos2016} and \cite{Pachet2018}).

We now move our analysis to a higher level, by exploring how artificial texts resemble the style of the training author. Stylistic traits are supposed to reflect subtle choices of the author in terms of vocabulary, syntactic constructions and structural composition, to mention a few. As such, a comprehensive quantification of the \emph{style} of an author is out of reach. On the other hand, a very simple feature such as the frequency distribution of $n$-gram of letters has been successfully selected as a key ingredient in some of the most effective approaches to authorship attribution~\cite{Basile2008,Benedetto2013}.

We use one of the state-of-the-art algorithms to test the automatic attribution of the author of our artificial texts. The implemented method is in fact one of the two methods that have been succesfully used for the attribution of Antonio Gramsci's papers~\cite{Basile2008}.
Essentially, each method defines a kind of \textit{similarity distance} between texts.
Let us very briefly describe just the first method used here, referring to~\cite{Benedetto2013} for further details.
The method is based on (characters) $n$-grams and it is probably one of the simplest possible measures on a text: after a first experiment based on bigram frequencies presented in 1976 by Bennett~\cite{Bennett1976}, Ke\u{s}elj et al~\cite{Kes03} published in 2003 a paper in which $n$-gram frequencies were used to
define a similarity distance between texts (see also~\cite{Clement2003}).
The similarity distance was introduced and discussed in~\cite{Basile2008}: we call $\omega$ an arbitrary $n$-gram, and we denote by $f_{X}(\omega)$ and $f_{Y}(\omega)$ the relative frequencies with which $\omega$ occurs in text $X$ and $Y$. $D_{n}(X)$ is the \textit{$n$-gram dictionary} of text $X$, that is, the set of all $n$-grams which have non-zero frequency in $X$ (similarly for $Y$) and we define what we will call the $n$-gram distance between text $X$ and text $Y$ as\footnote{To be more precise, $d_{n}$ is a pseudo-distance, since it does not satisfy the triangle inequality and it is not even positive definite: two texts $X,Y$ can be at distance $d_{n}(X,Y)=0$ without being the same.}:
\begin{equation}
d_{n}(X,Y):=\frac{1}{|Z|}\,\sum_{\omega\in D_{n}(X)\cup D_{n}(Y)}\left(\frac{f_{X}(\omega)-f_{Y}(\omega)}{f_{X}(\omega)+f_{Y}(\omega)}\right)^{2}\label{ngramdis}
\end{equation}
where the denominator $|Z| = \left\vert D_{n}(X)\right\vert + \left\vert D_{n}(Y)\right\vert$ is the sum of the number of different $n$-grams in the two dictionaries, respectively, while the inner sum is taken over all the different $n$-grams occurring in the two texts.

Suppose the goal is to decide whether a given document $X$ has been authored by author $A$ or not. The approach adopted in~\cite{Basile2008} consists in first collecting $m$ documents from author $A$ and $m$ documents from an author $B$ (or, in general, from more authors). The distance of the candidate text $X$ to these documents is then used, with the help of a simple probabilistic method, to produce a similarity index $I(X)$, $-1\leq I(X)\leq 1$ (see~\cite{Basile2008} for details on the method). Values of the index close to $1$ (respectively, $-1$) reveal a very strong attribution to author $A$ (respectively, $B$), while values close to 0 indicate a very weak attribution (see~\cite{Basile2008,Benedetto2013} for more details).

\section{Corpora}
\label{sec:corpora}

Our aim was to train an LSTM network on a large corpus obtained from a single author, in order to perform also the experiments on authorship attribution and to assess whether the model was capable of capturing some stylistic features of its ``teacher''. We employed the works by Charles Dickens, since he was a very prolific author whose bibliography is freely available in plain text format at ProjectGutenberg.\footnote{\url{http://gutenberg.org}}
We collected a total of eighteen works by Charles Dickens, which resulted in a corpus of over 24 millions characters.\footnote{We used the following works: A tale of two cities, David Copperfield, Oliver Twist, Bleak house, Great expectations, The life and adventures of Nicholas Nickleby, The old curiosity shop, The Pickwick papers, Dombey and son, Little Dorrit, Life and adventures of Martin Chuzzlewit, Our mutual friend, Barnaby Rudge, A Christmas carol, The uncommercial traveller, Hard times, Letters, A child's history of England.}

For the authorship attribution experiments, we also collected a smaller corpus of texts, some still from Charles Dickens, and some others from a different author. Clearly, these additional texts from Dickens needed to be disjoint from those in the larger corpus, on which the LSTM network had to be trained.\footnote{We used excerpts from the following works: Signal-man, A Christmas tree, The poor relation's story, The schoolboy's story, Hunted down, Pictures from Italy, The chimes, The haunted man and the ghost's bargain, Tom Tiddler's ground, The wreck of the Golden Mary.} Regarding non-Dickens documents, we collected texts by Robert Louis Stevenson, who was also a prolific author of the XIX century.\footnote{We used excerpts from the following works: Treasure island, The strange case of Doctor Jekyll and Mister Hyde, Kidnapped, The Black Arrow.} For this second corpus, we collected 30 documents both for Dickens and for Stevenson, each consisting of 10,000 characters.

\section{Experiments}
\label{sec:experiments}

The experiments with LSTMs were run using the \texttt{torch-rnn}
package.\footnote{\url{https://github.com/jcjohnson/torch-rnn}}
We trained an LSTM network with two layers and 1,024 cells in each layer. As
customary in text generation experiments with
LSTMs~\cite{Karpathy2015understanding}, the training set was split into chunks
of length equal to $K$ characters: in this way, gradients in truncated
backpropagation are propagated up to $K$ steps back, but the status of each LSTM
cell is not reset after each example, so that longer-range correlations can in
principle be learnt. We firstly present results obtained using $K=100$, later investigating the impact of such hyper-parameter on the considered evaluation metrics.
To avoid overfitting, a dropout equal to 0.7 was applied, and a validation set (4\% of the whole corpus) was exploited to
monitor the loss function during training.

To compare the statistics of the LSTM texts with that of other
non-trivial models, we used a plain $m$-order Markov process as a baseline. A family of $m$-order Markov models were trained from our corpus, with $m$ taking values 2, 4, 6, 8, 10, and 12. The models
were used to generate artificial texts starting from a seed taken from the
original corpus.

Tables~\ref{tab:text_samples} and~\ref{tab:text_markov} show some examples of
texts generated with different temperatures from the LSTM, and by different
Markov models, respectively.
The whole set of generated samples is publicly available at \url{http://agentgroup.unimore.it/Lippi/generated_samples_Dickens.zip}.

\begin{table*}
  \begin{center}
    \caption{Some samples generated by LSTM, as a function of the temperature
$T$.\label{tab:text_samples}}
  \begin{tabular}{|c|l|}
  \hline
  $T$ & LSTM generated text \\
  \hline
  \multirow{2}{*}{0.1} & I had no doubt that I had no doubt I had no sooner said
that I had no sooner said\\
  & that I had no sooner said that I was a stranger to me to see him\\
  \hline
  \multirow{2}{*}{0.3} & I will not see him as I have been the family of my heart and expense,\\
  & and I should have seen him so soon.\\
  \hline
  \multirow{2}{*}{0.5} & 'I am sure I think it is,' said the doctor, looking at
him at his heart on the window,\\
  & and set him there for a short time, 'that I shall find the girl there.\\
  \hline
  \multirow{2}{*}{0.6} & The old man entered the other side, and then ascended the key on his shoulder.\\
  & 'I think I have no doubt, sir,' replied the woman. \\
  \hline
  \multirow{2}{*}{0.7} & 'I can refer to the world,' said Mr. Tupman, suspiciously,\\
  & 'that the same brother was on the provoked passage.' \\
  \hline
  \multirow{2}{*}{0.8} & You look at me, you know I don’t think it should be what I have understood.\\
  & I know what she has of no confidence, and I have first cheered you.\\
  \hline
  \multirow{2}{*}{0.9} & And so the position and paper escaped you by the Major, with the neck,\\
  & by my own evening, that there was a shadow of it. \\
  \hline
  \multirow{2}{*}{1.0} & This interference of point that was always large in the evening, which was,\\
  & and used to save the crooked way, and could leave the undertakers upon it.\\
  \hline
  \multirow{2}{*}{1.2} & Softly. They rose and who faithfully as stalled off, and his distrust\\
  & in the tip of which power it was.  \\
  \hline
  \multirow{2}{*}{1.5} & As if us, she Immoodnished Mrs Jipe Town horsemaking,\\
  & like the nights foldans with mid-yoUge false. Half-up?\\
  \hline
  \multirow{2}{*}{2.0} & Connursing, visibly; brassbling, on what cohn;
or’Pertixwerkliss issuin'p).\\
  & haf-pihy; and he-carse masls anycori’; nod me: full, nor cur your two
yellmoteg'\\
  \hline
  \multirow{2}{*}{3.0} & 'Siceday; Quaky o’k, ,-GNIRRZVRIIMoklheHw, eab-mo,'\_ ventvedes.r.'\\
  & Egg\_iglazzro!’wM’. 'Sav nam. Ebb-\_Edjaevi." \\
  \hline
  \end{tabular}
  \end{center}
\end{table*}

\begin{table*}
  \begin{center}
    \caption{Some samples generated with Markov models of increasing
order.\label{tab:text_markov}}
  \begin{tabular}{|c|l|}
  \hline
  order & Markov generated text \\
  \hline
  \multirow{2}{*}{2} & 'And Mr Butentime ther foreweemair Masperf torto lit, 'It
make to to yesee!" \\
  & shis thed to goin blie, thave com of a dess at's mand havestroult frot own:
ady. Saint\\
  \hline
  \multirow{2}{*}{4} & But the Nation, and you in looking at all,' said Mrs.
Kenwigs's hospite as I have gover with now,\\
  & 'and the eyes. Perhaps, and even to help, ‘I am sure loving her,\\
  \hline
  \multirow{3}{*}{6} & Sam eyed Oliver, with some to her motion of the ladies,
which at no immense\\
  & mob gathering which the after all the adorned to nod it, Sir!' said
Fledgeby\\
  \hline
  \multirow{2}{*}{8} & The foot of the medical manner, "Jenny saw, asserting to
the large majority,\\
  &just outside as he had hardly achieved to be induced the room.\\
  \hline
  \multirow{2}{*}{10} & I endeavour to get so precious contents, handed
downstairs with pleasure the great Constitutions. \\
  & It happened to him like a man,  and the fallacy of human being accepted
lover of the tide\\
  \hline
  \multirow{2}{*}{12} & Mr Witherden the notary, too, regarded him; with what
seemed to bear reference to the friar,\\
  & taking this, as it served to divert his attention was diverted by any
artifice.\\
  \hline
  \end{tabular}
  \end{center}
\end{table*}

\begin{figure*}[!t]
\begin{center}
A)\includegraphics[width=0.47\textwidth]{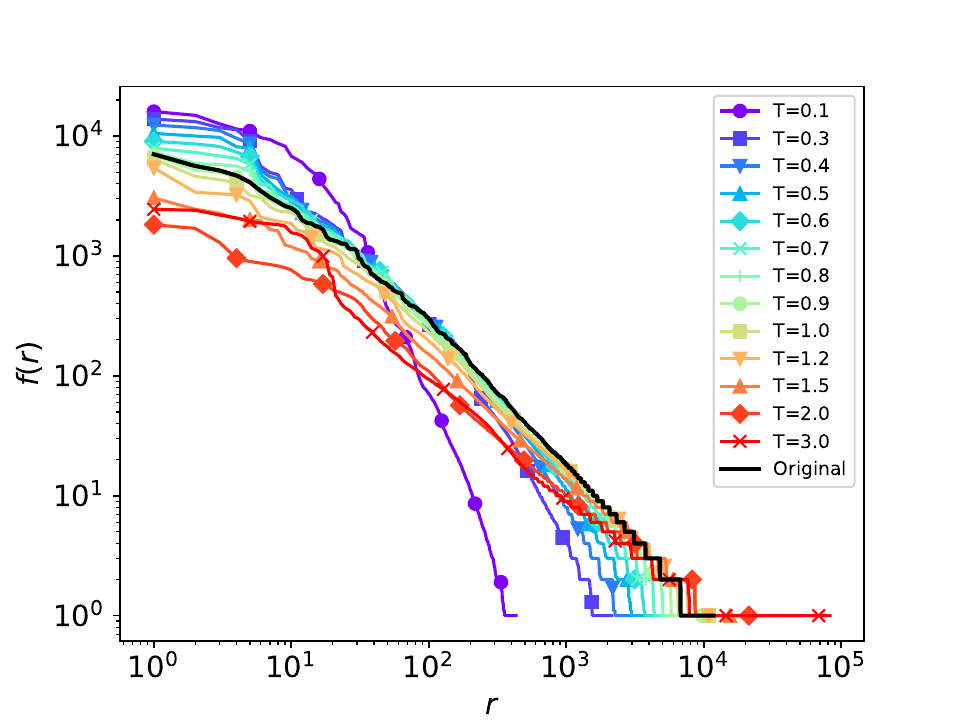}
B)\includegraphics[width=0.47\textwidth]{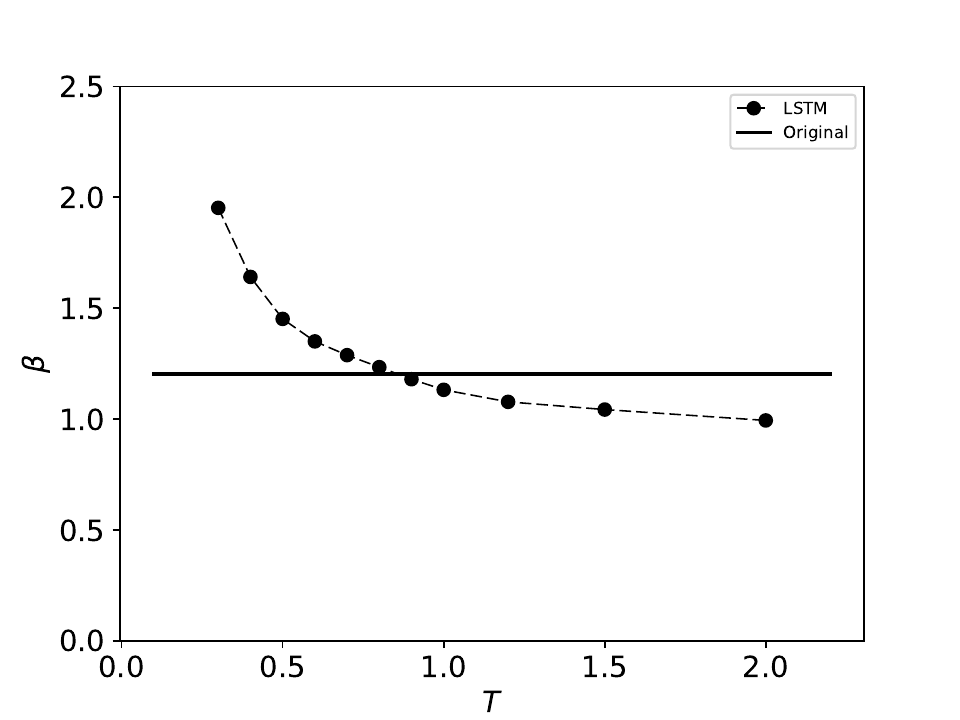}\\
C)\includegraphics[width=0.47\textwidth]{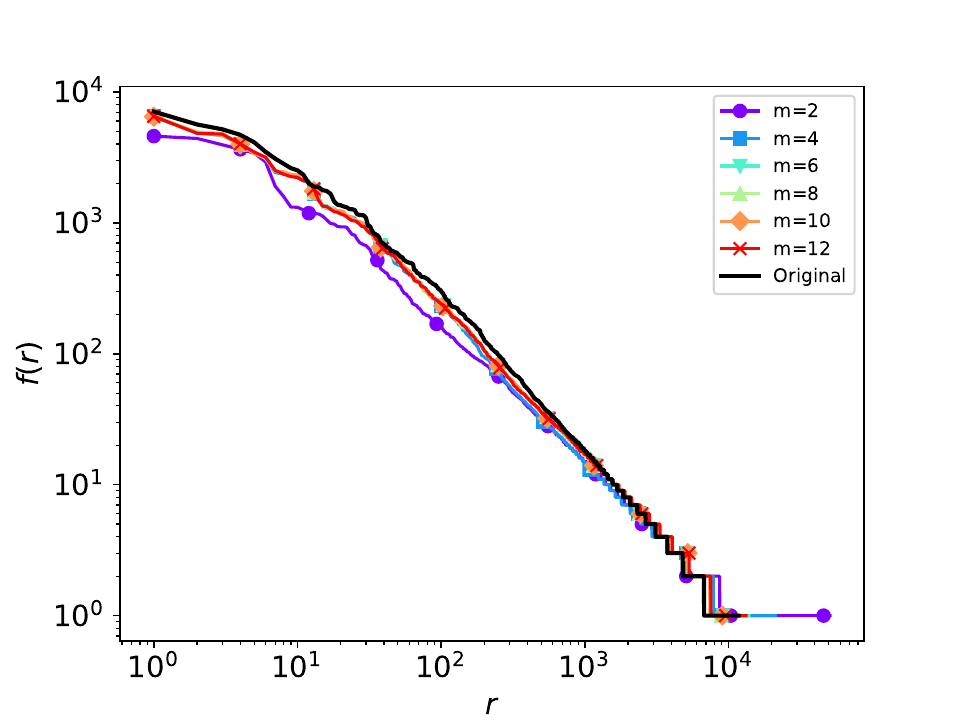}
D)\includegraphics[width=0.47\textwidth]{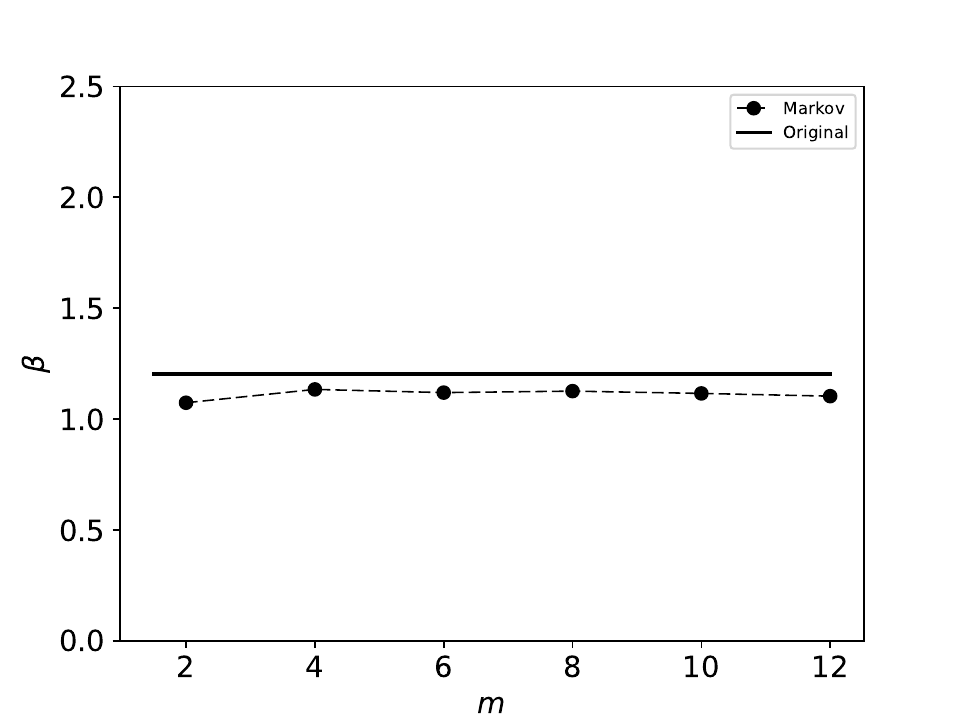}
\caption{{\bf Zipf's law}. A-C) Zipf's
rank-frequency distribution in LSTM (respectively, Markov) texts, with temperature $T$ in the range 0.1-3.0 (respectively, order $m$ in the range 2-12). B-D) Exponent $\beta$ measured in the region between ranks $10^2$ and $10^3$, as a function of $T$ (LSTM) or $m$ (Markov), with dashed line corresponding to the exponent measured in the original corpus.
For readability, panels A-C do not show markers for every point. Figure best seen in colors.\label{fig2}}
\end{center}
\end{figure*}

\begin{figure*}[!t]
\begin{center}
A)\includegraphics[width=0.47\textwidth]{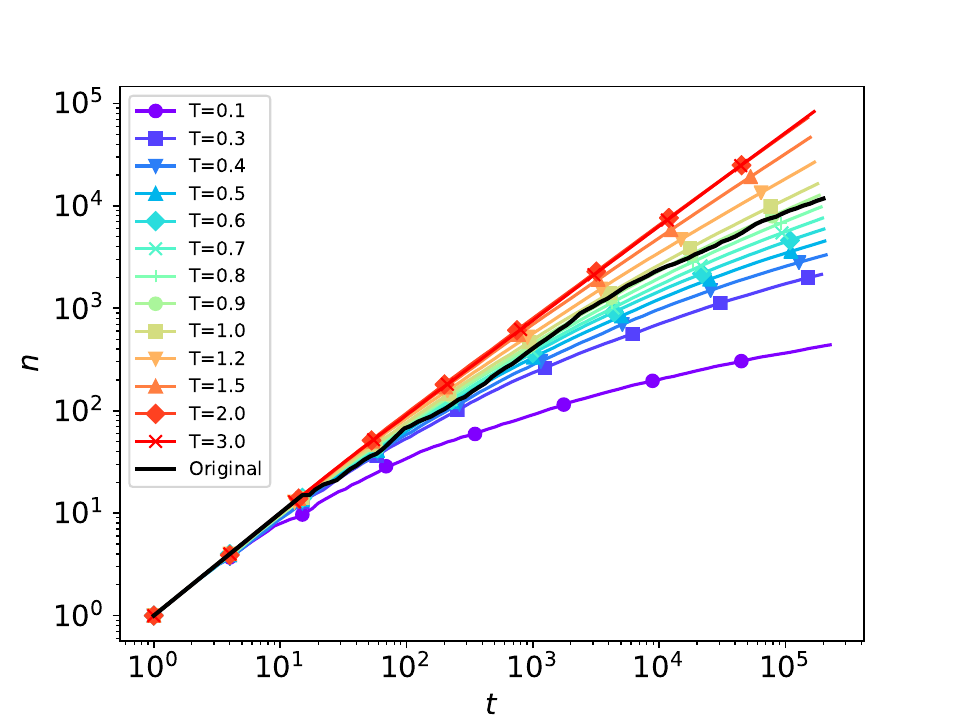}
B)\includegraphics[width=0.47\textwidth]{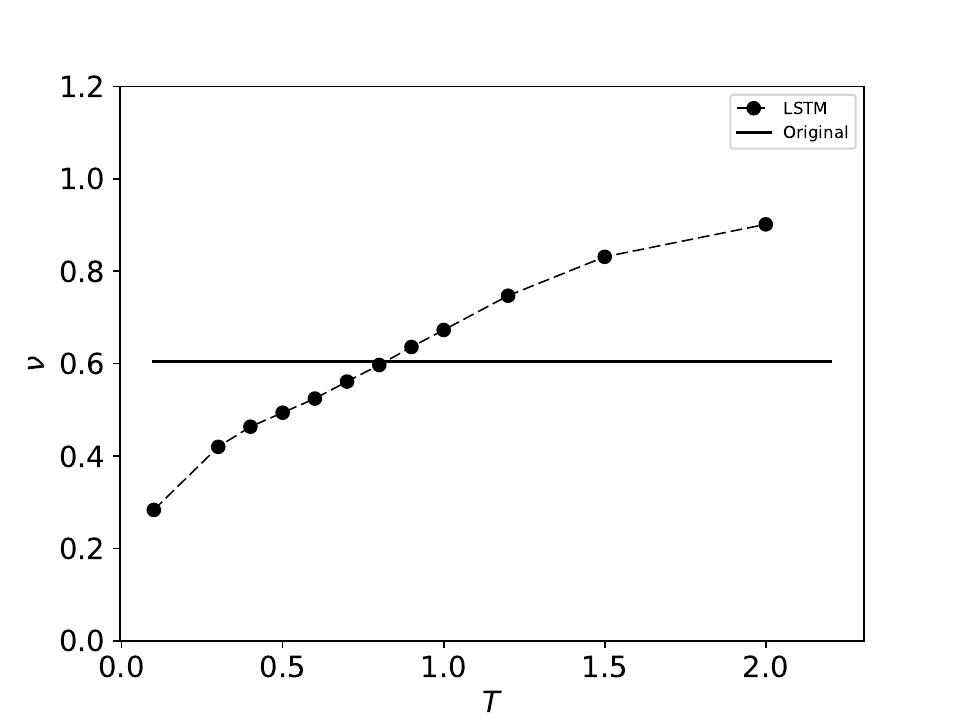}\\
C)\includegraphics[width=0.47\textwidth]{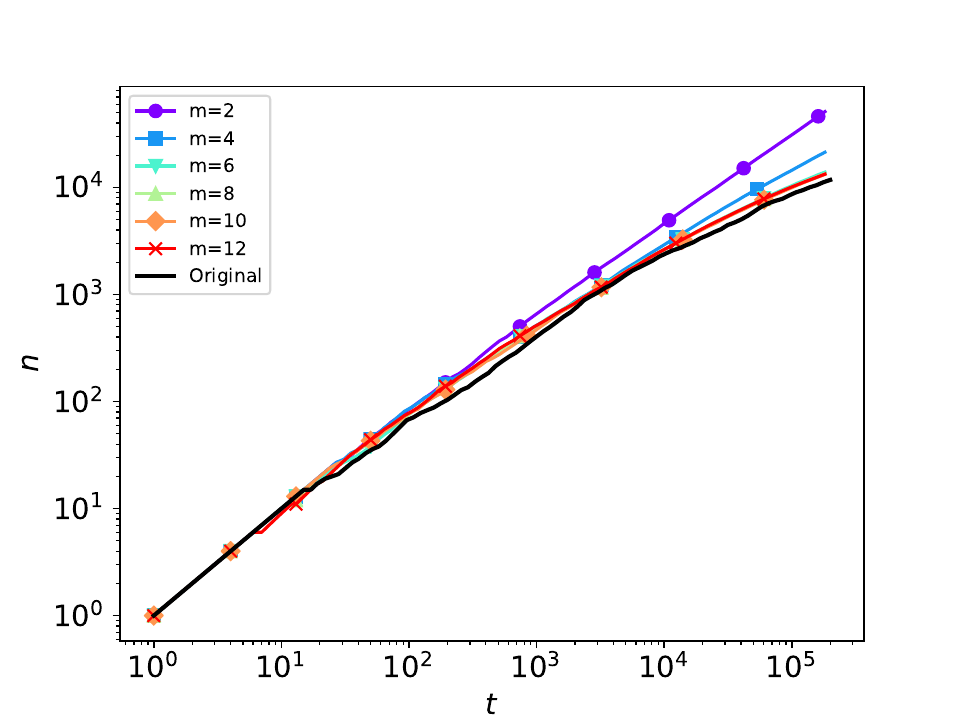}
D)\includegraphics[width=0.47\textwidth]{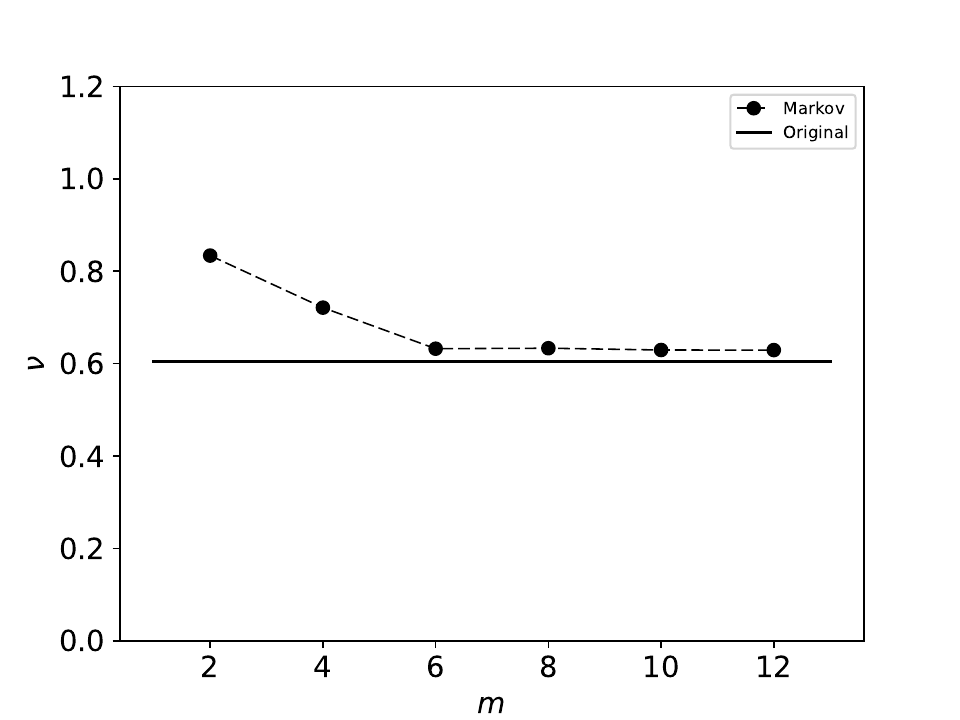}
\caption{{\bf Heaps' law}. A-C) Heaps' growth curve for the LSTM (respectively, Markov) texts, for temperatures in the range 0.1-3.0 (respectively, order in range 2-12). B-D) Exponent $\nu$ measured in the $t>1000$ region of
the Heaps's curve for LSTM (respectively, Markov) texts as a function of the temperature (respectively, order).
In panels B and D the dashed line corresponds to
the exponent for the Dickens' corpus. Figure best seen in colors.\label{fig3}}
\end{center}
\end{figure*}

\subsection{Zipf's and Heaps' Laws}
As a first test comparing the statistics of the LSTM-generated texts to other
models capable of rendering stochastic versions of texts, we looked and the
distribution of words frequencies.
We first evaluated the Zipf's law, by measuring the relationship between the
rank in the set of words, ordered by frequency, and word frequency itself.
Figure~\ref{fig2}A shows the rank-frequency distributions of
words in the LSTM texts for temperatures in the range 0.1-3.0. The
plot also shows in black the result for the original Dickens corpus. We fitted the value of the exponent within the
region between ranks $10^2$ and $10^3$ to determine more clearly the behavior
with temperature, which showed a clear power-law behaviour consistently across all
but the two extreme temperatures. Figure~\ref{fig2}B shows the resulting value of the Zipf's
exponent $\beta$ as a function of temperature in the range 0.3-2.0. LSTM texts
generated with low temperatures have a frequency rank distribution which decays
faster with rank. On the contrary, for higher
temperatures the distribution flattens, showing a smaller exponent $\beta$. The
dashed line in the figure shows the value of the exponent estimated
from the Dickens' corpus, which intercepts the LSTM results at a
temperature between 0.8 and 0.9, approximately.

A similar analysis was performed on the Markov-generated texts.
Figure~\ref{fig2}C shows the rank-frequency distribution for texts generated
with Markov
models. Interestingly, there is little
variation of the distribution as a function of Markov
order. This is also corroborated by the estimation of the exponent (Figure~\ref{fig2}D), which shows the exponent of Markov texts with a value
slightly below that of the original source.

In order to assess the validity of Heaps' law for the artificial language,
we computed the statistics of the vocabulary growth for the LSTM and Markov texts.
Figure~\ref{fig3}A shows the results for the LSTM texts for a
range of temperatures from 0.1 to 3.0. For comparison, the curve for
the original text is shown in black. Figure~\ref{fig2}B represents the results
of
fitting the power-law behavior in the region for $t>1000$, using the same range
of temperatures shown for the $\beta$ exponent in Figure~\ref{fig2}B. The dashed
line, which represents results for the original text, cuts the LSTM data
at a point between temperatures 0.8 and 0.9.

\begin{figure*}[!t]
\begin{center}
A)\includegraphics[width=0.47\textwidth]{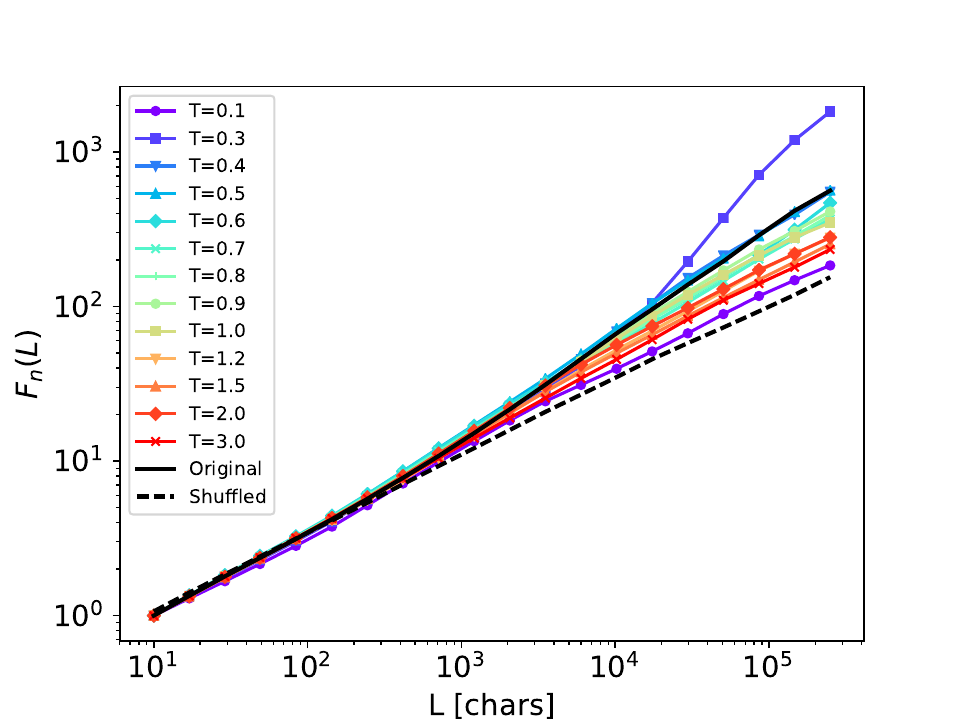}
B)\includegraphics[width=0.47\textwidth]{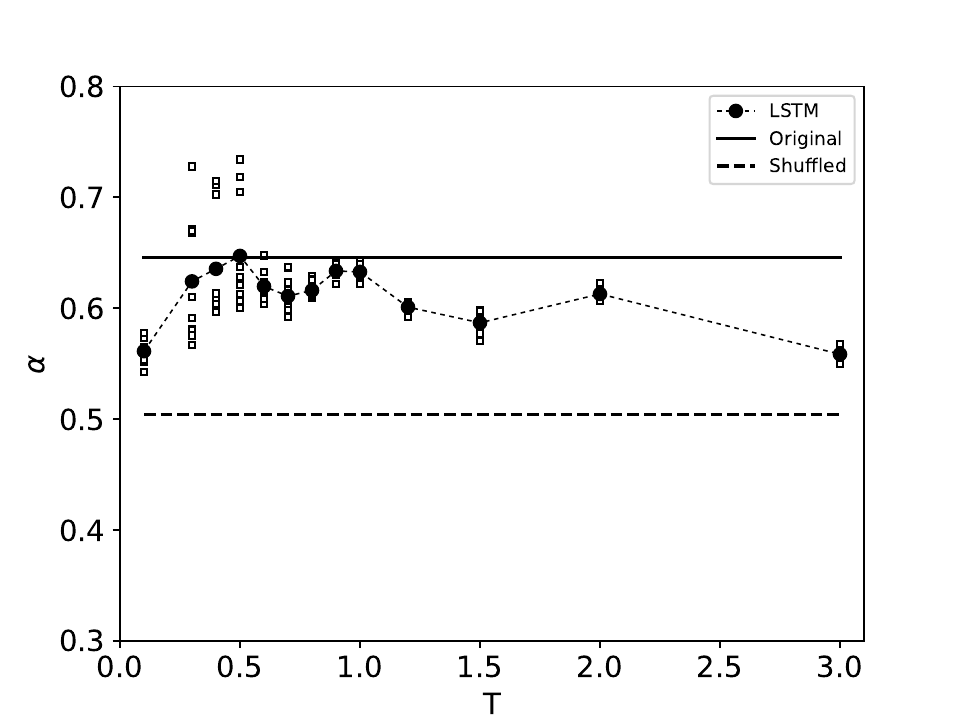}\\
C)\includegraphics[width=0.47\textwidth]{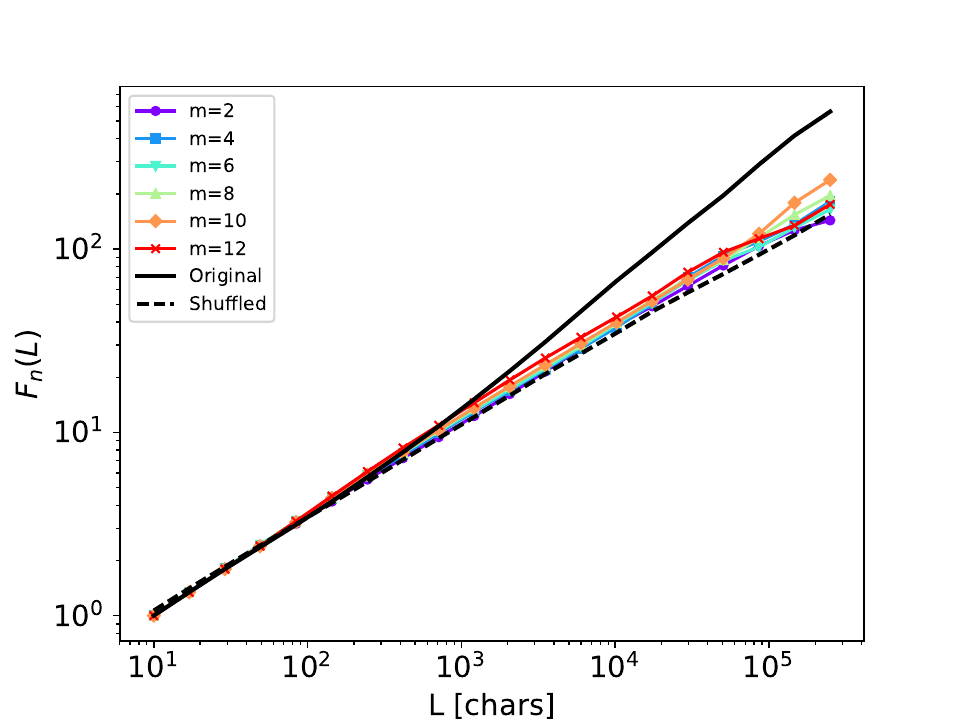}
D)\includegraphics[width=0.47\textwidth]{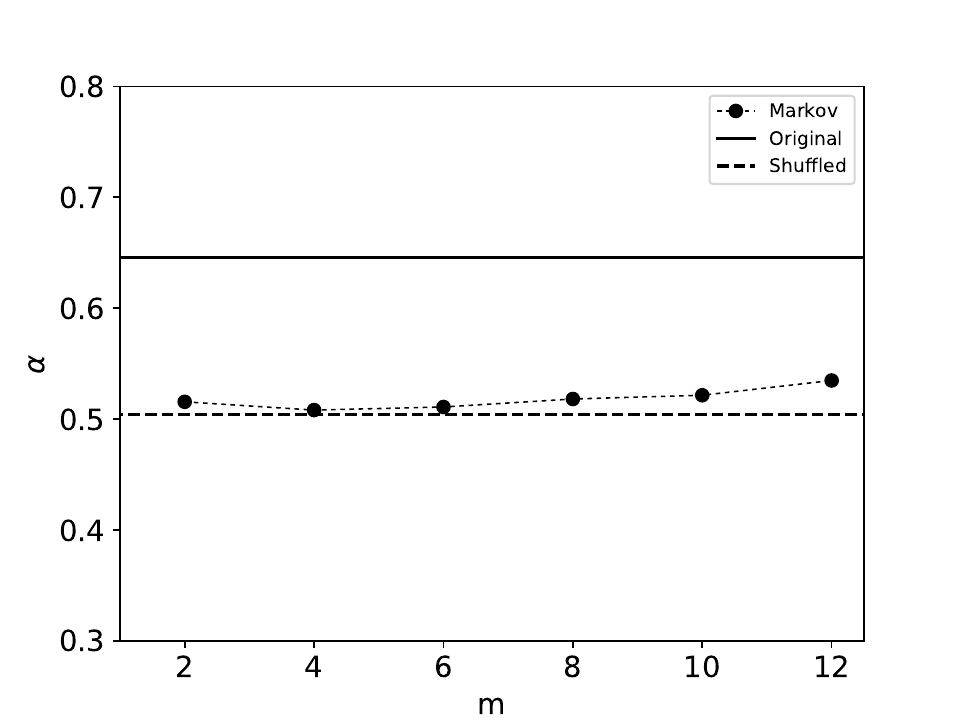}
 \caption{{\bf Long-range correlations}.
 A-C) Normalised fluctuations $F_n(L)$ as a function of the scale $L$
for LSTM (respectively, Markov) texts.
B-D)  Fluctuation exponent $\alpha$ obtained by fitting a power
law the LSTM and Markov data in panels A and C, in the range $L=[10^2,10^4]$. Black circles represent the mean over 10 samples, while small empty squares are the estimations for individual samples.
The full (dashed) black line represents the real (shuffled) text\label{fig4}.}
\end{center}
\end{figure*}

\begin{figure*}[!t]
\begin{center}
A)\includegraphics[width=0.47\textwidth]{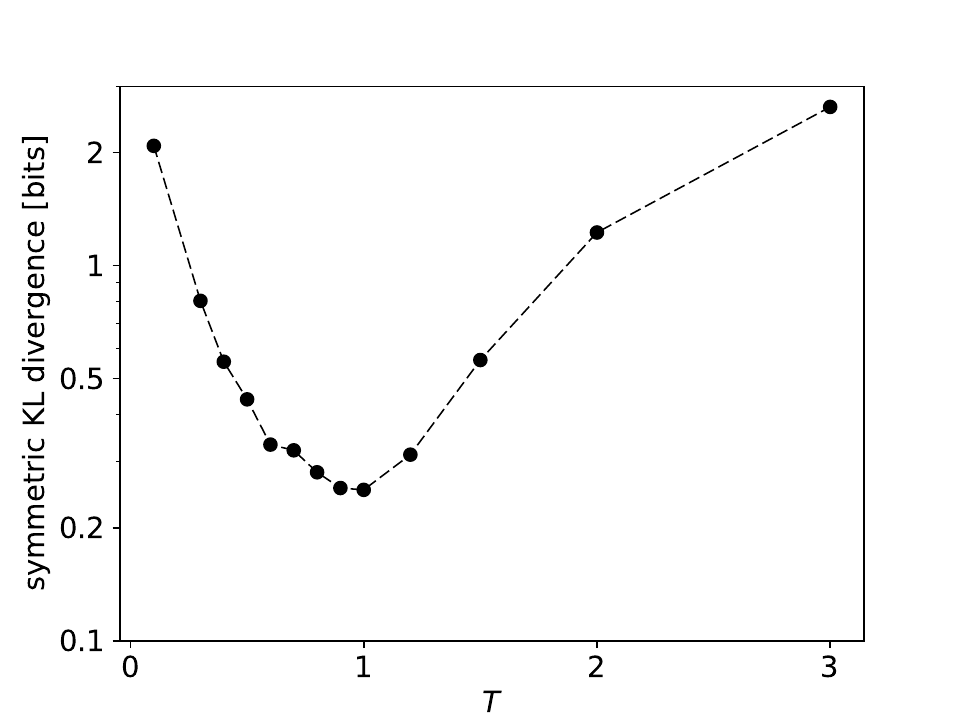}
B)\includegraphics[width=0.47\textwidth]{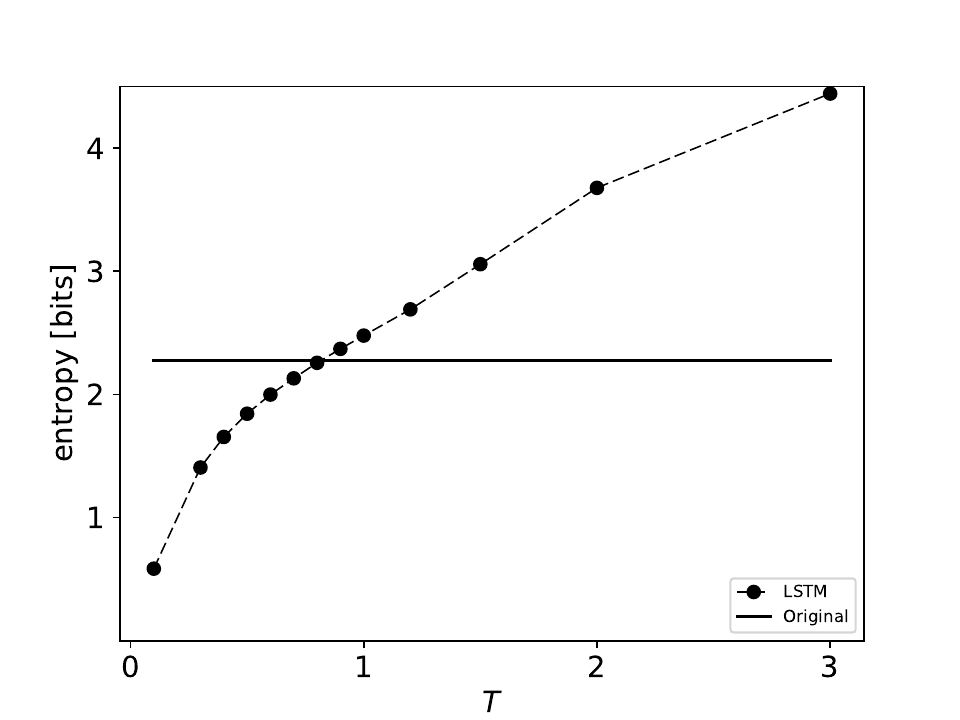}\\
C)\includegraphics[width=0.47\textwidth]{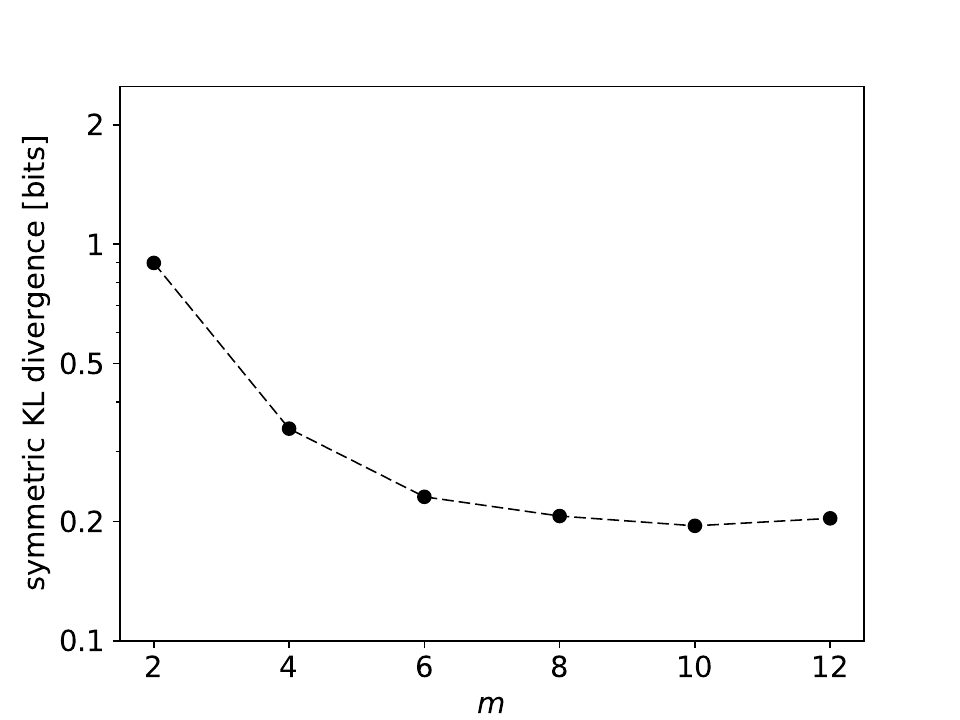}
D)\includegraphics[width=0.47\textwidth]{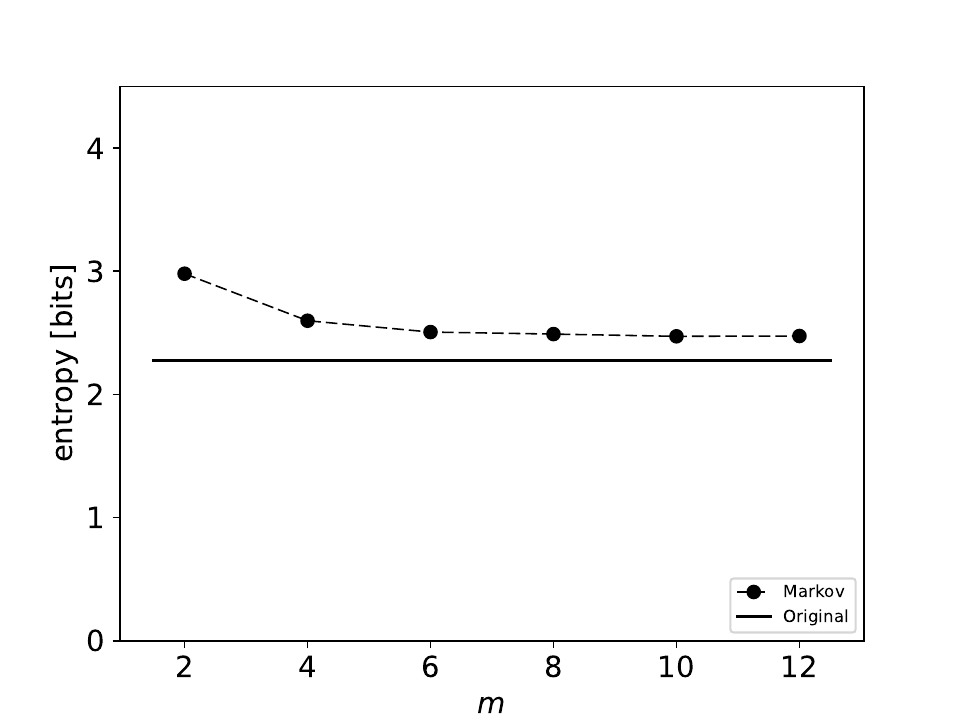}
 \caption{{\bf Divergence and entropy}.
Symmetrised KL-divergence and entropy, as a function of temperature for the
LSTM texts (panels A-B), and as a function of the order for the
Markov-generated texts (panels C-D). In panels B and D the solid line represents the entropy of the
original text.
\label{fig5}}
\end{center}
\end{figure*}

\subsection{Long-Range Correlations}
Beyond the statistics of word frequencies, natural texts show correlations that
span hundreds or thousands of characters, showing power-law decay. While a
direct measure of these correlations is possible in principle, more efficient
methods are based on spectral~\cite{Voss1975} or
fluctuation~\cite{Peng1992,Buldyrev1995} properties of the sequences. In
particular, we tested the LSTM generated texts
using Detrended Fluctuation Analysis (DFA) \cite{Peng1992,Buldyrev1995} applied
to linguistic character sequences.
By means of DFA it is possible to estimate such exponent
$\alpha$, by fitting a power law to the the dependence of the
fluctuations $F(L)$  with the scale $L$.  We estimated the dependence of $F(L)$
with the scale $L$ for LSTM texts generated at different temperatures
and for the Markov-generated texts for different orders.  Figure~\ref{fig4}A
shows the
normalised fluctuations $F_n(L)$ as a function of the scale $L$ for
LSTM-generated texts
together with the results for the original Dickens' corpus and a shuffled
realisation of it.
In all cases there are signs of power-law behavior for a wide range of scales,
with the exception
of the two extreme temperatures (0.1 and 3.0). Although the data
corresponding to the
shuffled text seems to have a less steep slope than the LSTM samples, a more
quantitative analysis is required
to compare the extent of correlations. Figure~\ref{fig4}B shows the
estimations of the exponent $\alpha$
obtained by fitting a power-law to the data in panel A in the region spanned by
scales $L=10^2\ldots 10^4$. For comparison,
the full black line corresponds to the result for the original
Dickens' text while the dashed line is the result obtained
from the shuffled text. The full black circles are the average of the
estimation of the exponent from the 10 available samples,
whereas small squares show the result for each sample. It is in the
region between of temperatures between
0.5 and 1.0 that the exponent $\alpha$ es closet to the value for the original
text. Yet, while the dispersion of individual
sample measures are very broadly spread for temperatures close to 0.5 they are
tightly clustered around
the mean for $T=1.0$. A similar analysis was done on the Markov-generated
texts. Figure~\ref{fig4}C shows the result of the DFA for Markov texts
of orders 2--12, and a comparison to the original and shuffled texts:
clearly, all the Markov-generated texts have a
structural correlation that resembles more the shuffled text than the
original one. This trend is corroborated in Figure~\ref{fig4}D, showing
the estimated
exponent $\alpha$ as a function of Markov order. In all cases,
such value is very close to that obtained by the shuffled
text.

\subsection{Entropy}
Our final test to probe the statistical structure of the LSTM texts consisted in
the estimation of entropy measures. The first estimation
corresponded to a symmetrized KL-divergence (see Section~\ref{sec:methods})
between the LSTM for different temperatures and original text, estimated by
means of compression algorithms that are sensitive
to the long-range structure of the signal. Figure~\ref{fig5}A shows the
divergence as a function of $T$. There is a
 well-defined minimum at $T\sim1$, indicating that at that temperature the
structure of the LSTM text is closest to the original.
Figure~\ref{fig5}C shows
 a similar plot for Markov texts, which also
approximate the structure of the original texts asymptotically for larger
orders.
While the previous analyses showed the behavior of a distance-like quantity,
the entropy is a more direct estimation of the statistical structure of a
signal. Panels B and D in Figure~\ref{fig5} show the value of the entropy for
the
LSTM and Markov texts as a function of the temperature and order respectively.
In both cases the solid line corresponds to the value of the entropy of the
original text. For LSTM, lower $T$ values produce texts with small
entropy, thus easier to compress, since they show repetitive patterns. As $T$ grows, entropy also grows, with the texts becoming the most
similar to the original around $T\sim0.9$ (in line with our analyses) and
far more disordered for larger $T$ values.
Markov texts, instead, show a monotonic approximation to the entropy of the
original text, slightly above even for higher orders.

To complete the analysis, we also computed the perplexity of LSTM-generated texts, using the \texttt{KenLM}
library\footnote{\url{https://github.com/kpu/kenlm}}. To do this, we learned a
bigram language model on the original Dickens corpus and then we computed the
perplexity both for the artificial texts,
and for the off-sample Dickens documents in the test corpus used for authorship
attribution. Results are perfectly in line with those obtained with
entropy computation via compression, with values of temperature around 0.9-1.0
producing texts that are the most similar to the original.

\begin{figure*}[!t]
\begin{center}
A)\includegraphics[width=0.47\textwidth]{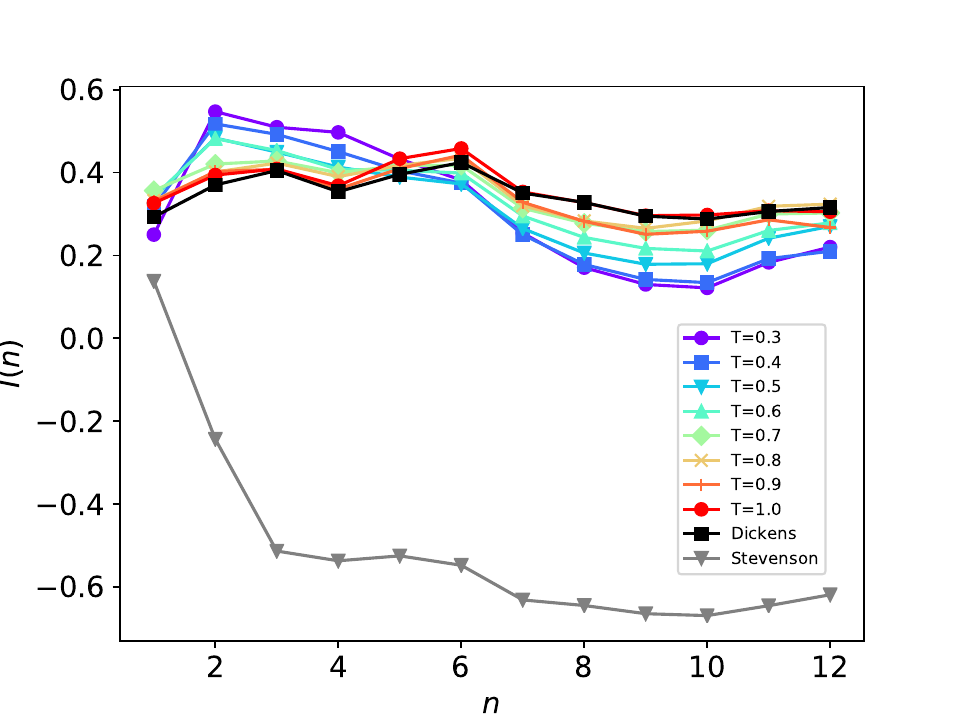}
B)\includegraphics[width=0.47\textwidth]{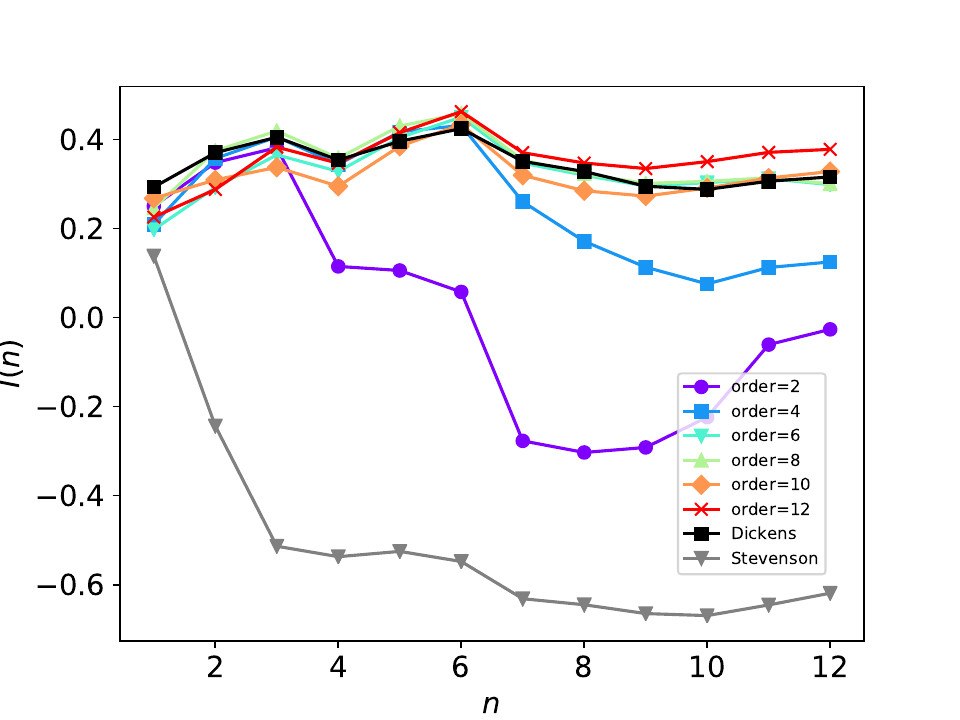}\\
 \caption{\textbf{Authorship attribution} for LSTM (A)
 and Markov texts (B). We show the attribution index
$I(X)$ for a text $X$ as a function of $n$-grams and as a function of $T$ (for LSTM) and $m$ (for Markov models). Indices closer to 1 (respectively,
-1) indicate stronger attributions to Dickens (respectively, Stevenson).\label{fig:aa}}
\end{center}
\end{figure*}

\subsection{Creativity and Authorship Attribution}\label{sec:exp_creativity}

To assess the degree of plagiarism (and, thus, of creativity) of our models, we measured the length of the Longest Common Subsequence
(LCS) between the artificial texts and the training corpus. As for Markov texts, not surprisingly the length of the LCS rapidly grows
with the order of the model, from a value of 19 for order 2, up to 73 and
125 for order 10 and 12, respectively. For LSTMs, the length of the LCS
instead results to be lower, with values comprised between 40 and 60 when $T$ is in the range between 0.4 and 1.2. For smaller and larger
temperatures, the LCS is clearly shorter, as the generated texts
have a less realistic structure, thus it is less probable to encounter
patterns identical to the original.
As a comparison with what we could call a sort of self-plagiarism,
encountered in a subset of the training corpus, we also computed
the length of the LCS of the novel ``Oliver Twist'' with respect to five other novels by Dickens (David Copperfield,
A Tale of Two Cities, Bleak House, Great Expectations, Hard Times): the LCS in this case resulted to be 45 characters long.

For the experiments on authorship attribution, we used the Dickens and Stevenson
corpora described in Section~\ref{sec:corpora}. For a fixed temperature of the
generated text, attribution is performed by averaging over $10$ samples of
artificial texts, using $n$-grams from $n=1$ up to $n=12$. 
To assess the validity of the attribution algorithm, we
compare the attribution of the generated texts with that of real texts. In
particular, for each of the 30 Dickens and Stevenson documents, we employ the
remaining 29 documents of each group to perform attribution to either author. Figure~\ref{fig:aa} shows the value of the index $I(X)$ defined
in Section~\ref{sec:creativity} as a function of $n$-grams (on the x-axis) and
either temperature for LSTM (panel A) or order for Markov texts (panel B), respectively.
The attribution of real texts performs extremely well, with a maximum for $n=6$ and
$n=10$ for Dickens and Stevenson, respectively. For LSTM, while the texts are indeed always
correctly attributed to Dickens, it is interesting to notice how the best
results are again achieved for values of $T$ around 0.8-1.0, while for
lower temperatures the attribution with larger $n$-grams drops,
as the generated texts tend to reproduce periodic (hence, less realistic)
patterns. Not surprisingly, Markov texts with $m \ge 6$ are also well attributed
to Dickens, due to the high degree of plagiarism described in Section~\ref{sec:exp_creativity}.

\subsection{Impact of $K$ hyper-parameter}
\begin{figure}[!t]
\begin{center}
\includegraphics[width=0.49\textwidth]{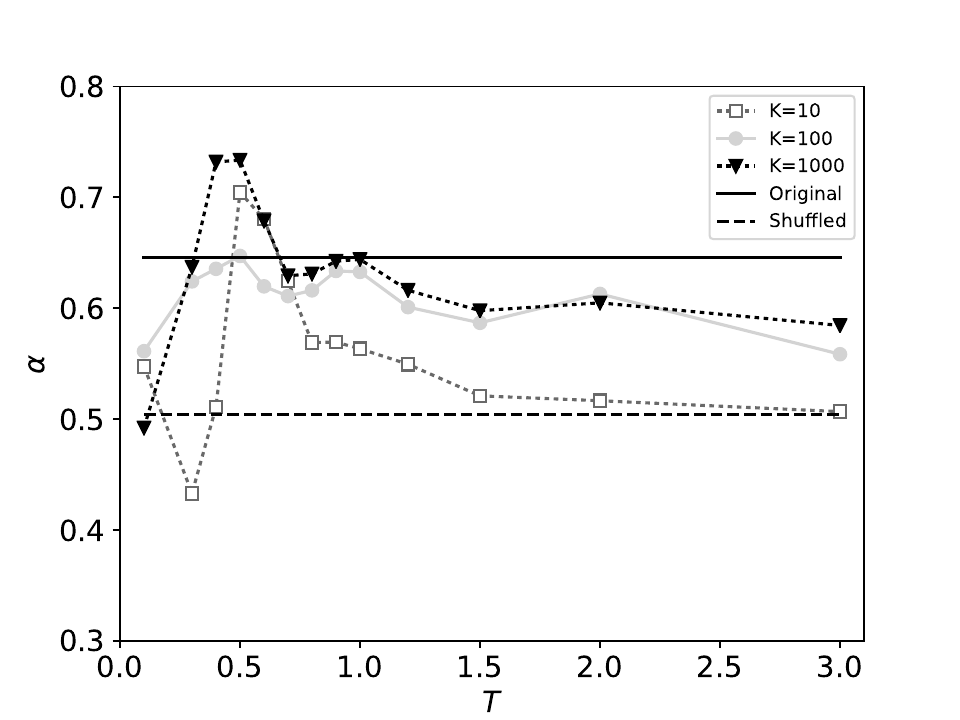}
 \caption{Exponent of DFA as a function of $T$, for different values of $K$.
\label{fig:comparison}}
\end{center}
\end{figure}

The results presented so far consistently show that there is a narrow range of temperatures around $T=1$ for which
the texts generated by LSTM are the most similar to the original, in terms of all the considered metrics.
We finally investigated the impact of $K$ hyper-parameter on the generated texts. Given that $K$ is the number of backward gradient propagation steps in the BPTT algorithm, it directly affects the way in which long-range information is propagated through the cells. Therefore, we considered two additional LSTM networks, trained with $K=10$ and $K=1,000$, respectively, and identical to the first network for all the other settings.
For all the considered metrics apart from long-range correlations, results are not significantly different from those obtained for the network trained with $K=100$ (see supplementary material for details). Long-range correlations are instead more significantly affected by the variation in $K$. From Figure~\ref{fig:comparison} we observe two different phenomena: (i) a peak for low-level temperatures, likely due to spurious, periodic patterns, that is visible also for $K=100$ (also notice the large variance across samples shown in Figure~\ref{fig4}); (ii) a dramatic drop in the exponent $\alpha$ for $K=10$ starting from $T=0.7$, whereas for $K=1,000$ the exponent is much closer to that of real text.

\section{Conclusions}
\label{sec:conclusions}

In this paper, we investigated at what degree texts generated by an LSTM network resemble texts generated by humans. To this aim, we presented an extensive experimental evaluation where we compared several characteristics of artificial and original texts, starting from statistical properties typically shown by natural language, such as the distribution of word frequencies and long-range correlations, up to higher-level analyses, such as the attribution of authorship.
Our study shows that LSTM-generated texts share key statistical features with natural language. In particular, the experimental results highlight the crucial role of the temperature parameter in producing texts that resemble those created by humans in their statistical structure, with an optimal range of temperatures, around $T=1$, that induce the highest degree of similarity. Interestingly, we also illustrated how a network trained on a single-author corpus can produce texts that are attributed to that author, according to authorship attribution algorithms.

The presented study suggests many interesting research directions, which we plan to investigate in future works.

First, we aim to compare the semantic information of original and artificial texts. It is clear from the samples shown in the experimental section, that LSTM texts are still far from human-generated texts in terms of meaningfulness, although showing similar statistical properties. It is thus possible that certain correlates of semantic information, like burstiness and clustering of keywords, are reflected in LSTM texts. Given that even the origin of long-range correlations in natural language is still debated, our work paves the way to deeper future investigations in this direction.

Secondly, we aim to extend the analysis of these statistical properties of the LSTM texts to different languages, in order to assess whether there are some languages that are easier or more difficult to reproduce for a machine.

Finally, we aim to extend the study on authorship attribution and plagiarism, for which in this paper we only presented preliminary results. For that purpose we plan to employ a larger corpus, in order to compare several authors, genres and languages, and to test different algorithms, as for instance those based on trainable machine learning systems~\cite{Neme2015,Jockers2010}.

\bibliographystyle{IEEEtran}
\bibliography{biblio}

%

\begin{IEEEbiography}[{\includegraphics[width=1in,height=1.25in,clip,
keepaspectratio]{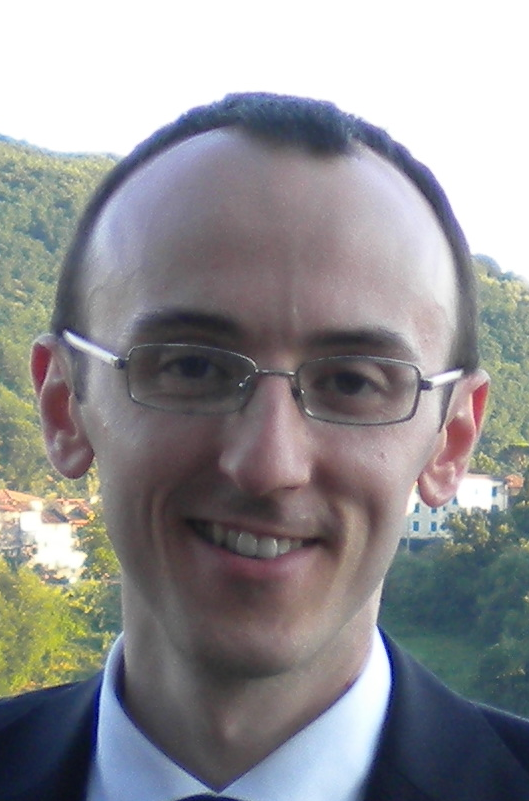}}]{Marco Lippi} received the PhD in Computer
and Automation Engineering from the University of Florence in 2010.
Currently, he is Assistant Professor at the Department of Sciences and
Methods for Engineering, University of Modena and Reggio Emilia. He previously
held positions at the Universities of Florence, Siena and Bologna, and he was
visiting scholar at UPMC, Paris.
His work focuses on machine learning and artificial intelligence, with
applications in bioinformatics, law, game playing, natural
language processing, and argumentation mining.
\end{IEEEbiography}

\begin{IEEEbiography}[{\includegraphics[width=1in,height=1.25in,clip,
keepaspectratio]{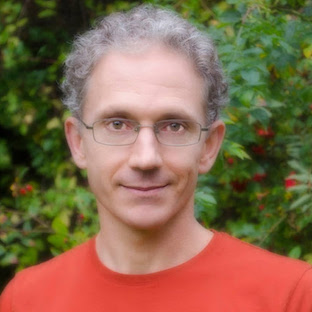}}]{Marcelo A. Montemurro} obtained a
PhD in Theoretical Physics in 2002 at the National University of C\'ordoba
(Argentina). He then moved to Italy with a fellowship from UNESCO to work at the
International Centre for Theoretical Physics in Trieste (Italy). In 2004 he
moved to the University of Manchester (UK) where he held a number of fellowships
before being appointed Lecturer. His research interests include the statistical
physics of complex systems and computational neuroscience.
\end{IEEEbiography}

\begin{IEEEbiography}[{\includegraphics[width=1in,height=1.25in,clip,
keepaspectratio]{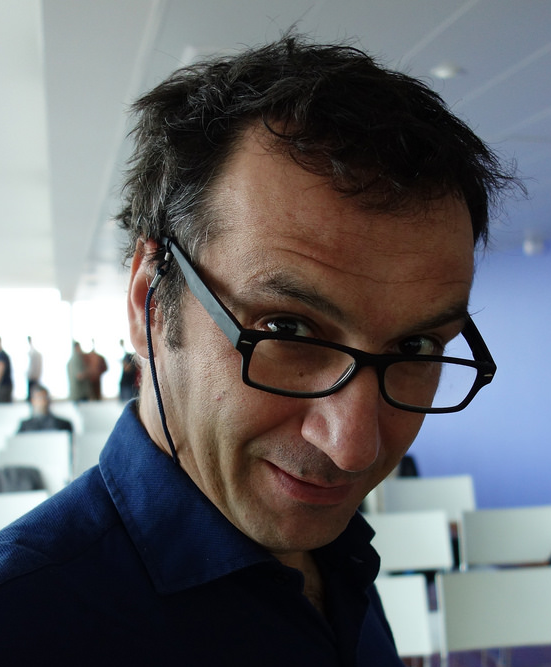}}]{Mirko Degli Esposti} is Full
Professor in Mathematical Physics in the Computer Science Department
at University of Bologna. He holds a Degree
in Physics and a PhD in Mathematics (Caltech and
PennState, USA). Lately he turned his
attention to applications of the theory of dynamical systems and of information theory to life science and human science, in particular human language.
\end{IEEEbiography}

\begin{IEEEbiography}[{\includegraphics[width=1in,height=1.25in,clip,
keepaspectratio]{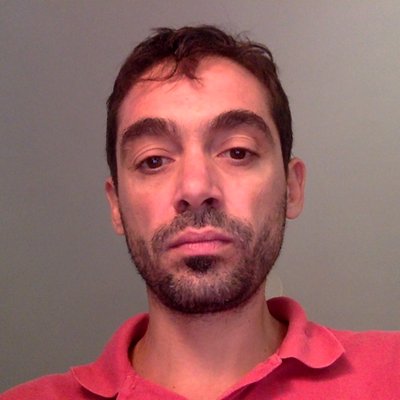}}]{Giampaolo Cristadoro}
is Associate Professor in Mathematical Physics at the
Department of Mathematics and Applications, University of
Milano-Bicocca. He received the PhD in Theoretical Physics from the University
of Insubria, Como and the University Paul-Sabatier, Toulouse.
He was postdoctoral fellow at the Max Planck Institute for the Physics of Complex Systems in Dresden, at the Center for Nonlinear and Complex Systems in Como, and at the Department of Mathematics at University of Bologna, where he became Assistant and Associate Professor.
His research interests include dynamical systems, probability and information theory, with applications to life science and natural language.
\end{IEEEbiography}

\end{document}